\documentclass{article}

\usepackage[final,nonatbib]{nips_2017}

\usepackage{graphicx}
\usepackage{subcaption}
\usepackage{cite}

\usepackage[utf8]{inputenc} 
\usepackage[T1]{fontenc}    
\usepackage{hyperref}       
\usepackage{url}            
\usepackage{booktabs}       
\usepackage{amsfonts}       
\usepackage{nicefrac}       
\usepackage{microtype}      

\title{Propagating Uncertainty in Multi-Stage Bayesian Convolutional Neural Networks with Application to Pulmonary Nodule Detection}

\author{
  Onur Ozdemir,\, Benjamin Woodward,\, Andrew A.~Berlin \\
  Draper\\
  \texttt{\{oozdemir,bwoodward,aberlin\}@draper.com} \vspace{-.4cm}
}

\begin{document}

\maketitle

\begin{abstract}
	Motivated by the problem of computer-aided detection (CAD) of pulmonary nodules, we introduce methods to propagate and fuse uncertainty information in a multi-stage Bayesian convolutional neural network (CNN) architecture. The question we seek to answer is ``can we take advantage of the model uncertainty provided by one deep learning model to improve the performance of the subsequent deep learning models and ultimately of the overall performance in a multi-stage Bayesian deep learning architecture?''. Our experiments show that propagating uncertainty through the pipeline enables us to improve the overall performance in terms of both final prediction accuracy and model confidence.
\end{abstract}

\section{Introduction  \& Related Work}\label{sec:bayesnet}


We introduce techniques for fusing model confidence information with pixel-level image information to enable uncertainty estimates to propagate, along with the image data, in a multi-stage Bayesian deep learning framework. Using computer-aided detection (CAD) of pulmonary nodules as an example, we demonstrate that Bayesian modeling in a multi-stage setting not only provides model confidence associated with nodule detection decisions, but, via propagation of uncertainty information between networks, also improves the overall detection/classification performance. This is important to build physician trust in the model's performance for applications such as cancer detection, where knowing that a result has high vs. low confidence can influence a follow-on treatment decision. To the best of our knowledge, none of the existing methods for lung lesion detection \cite{ypsilantis_arxiv16, setio_mi16, roth_mi16, zhu_arxiv17, ding_arxiv17} utilize or produce uncertainty/confidence information associated with nodule detection decisions.


The problem of computer-aided detection (or CAD) of pulmonary nodules using low-dose CT scans has a number of unique challenges \cite{anode09, luna16}. First, each scan is very large, prohibiting them from being used as full size 3D images for deep learning due to computational resource constraints. Second, pulmonary nodules to be detected are much smaller than full CT scans and they have high variability in terms of shape, size, and texture properties. These challenges have motivated researchers to divide the original problem into simpler subproblems and use multi-stage learning algorithms, where each algorithm attempts to solve a simpler subproblem. More specifically for the CAD problem, it is common to first segment 2D slices to find regions of interest (narrowing the search space) followed by performing 2D or 3D nodule detection within each region of interest to improve detection results.

\begin{figure}[t!]
	\centering
	\includegraphics[width=1\linewidth]{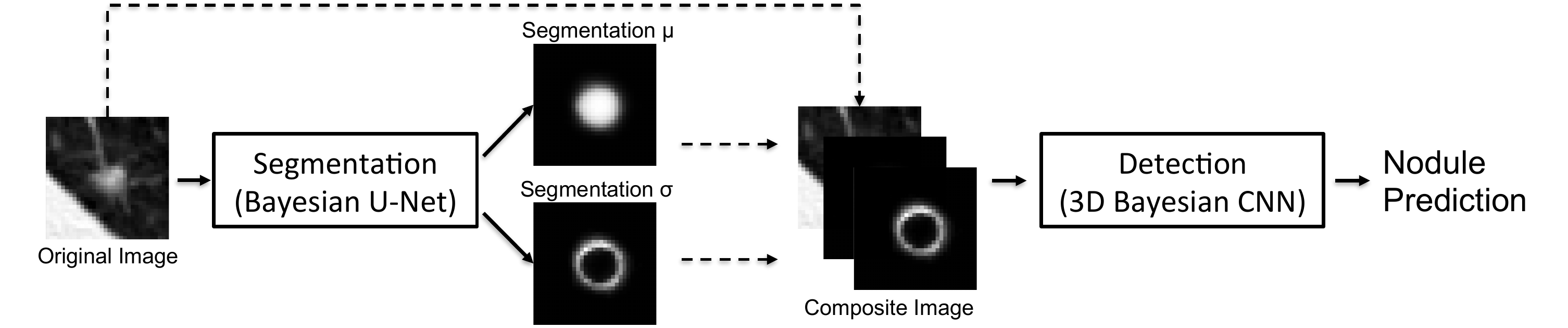}
	\caption{Two-Stage Bayesian CNN CAD architecture. Segmentation predictive mean and standard deviation maps are fused with the original image to form a 3-channel composite image, which is then fed into a 3D Bayesian CNN for final nodule detection. } \vspace{-.25cm}\label{fig:arch}
	
\end{figure}

Motivated by the CAD problem, we consider a multi-stage deep learning architecture as shown in Figure \ref{fig:arch}, comprising two consecutive Bayesian convolutional neural networks (CNNs) cascaded in a way that the predictions of the segmentation network inform the predictions of the nodule detection network. The overall performance of such an architecture depends on the individual performance of each network as well as on the coupling between them. In this architecture, each network has a different local view of the full CT scan. The segmentation network operates on full 2D axial CT slices (see Figure \ref{fig:seg_ex_orig}), whereas the detection network operates on small 3D volumes. 

Our goal here is to use segmentation predictions along with their uncertainties to improve nodule detection performance. To that end, segmentation predictions are incorporated as additional features for nodule detection by concatenating/fusing segmentation predictive mean and standard deviation with the original image to form a 3-channel composite image, as shown in Figure \ref{fig:arch}. This composite image is then fed into a Bayesian nodule detection network. The extra information provided by segmentation predictions along with associated uncertainties not only improves the detection performance for nodules that the segmentation network has high confidence of, but it also increases the overall model confidence of the nodule detection network. Since the segmentation predictions are based on 2D local context and as a result can produce false positives and lower confidence true positives, we also use the original 1-channel image to train a second Bayesian nodule detection network. Combining predictions from these two detection networks enables us to combine the strengths of both approaches, resulting in improved overall accuracy and higher model confidence. Although the specific application we consider here is pulmonary nodule detection, the approach of propagating and fusing uncertainty can be used in other multi-stage deep learning applications such as autonomous driving,  where a first network detects and classifies objects in the scene followed by a second network that makes decisions as to which direction to steer the car.     

 

\section{Two-Stage Bayesian CNN CAD Architecture \& Uncertainty Fusion}

As described in the previous section and shown in Figure \ref{fig:arch}, we have two Bayesian CNNs in our CAD architecture. Each network has a different local view of the full CT scan. 
For 2D segmentation (1st stage of the CAD architecture), we employ a 10-layer U-net architecture \cite{unet_miccai15} as a base network and add dropout with probability of $0.5$ after each convolutional layer. Using stochastic dropout at test time \cite{gal_icml16, gal_iclr16}, we obtain $50$ Monte Carlo (MC) samples approximating the segmentation prediction probability distribution \cite{Kendall2015BayesianSM}. We note that, to our knowledge, Bayesian U-Net has not been employed for medical image segmentation before.

Segmentation alone is not sufficient for nodule detection as it results in a large number of false positives (see Section \ref{sec:res} for numerical results). To perform nodule detection with reduced false positive rates, we employ a 3D CNN (2nd stage of the CAD architecture) with three convolutional and two fully connected layers. This detection network operates on small 3D nodule candidates that are extracted based on stacked segmentation predictions from the 2D axial slices. The reason we use 3D volumes for nodule detection is that suspicious nodules have unique 3D features that differentiate them from normal lung lesions. Similar to the approach for segmentation, we perform approximate Bayesian inference via stochastic dropout at test time to obtain $50$ MC samples approximating the nodule prediction probability distribution. 

\begin{figure}[h!]
	\centering
	\begin{subfigure}[b]{0.49\textwidth}
		\includegraphics[width=\textwidth]{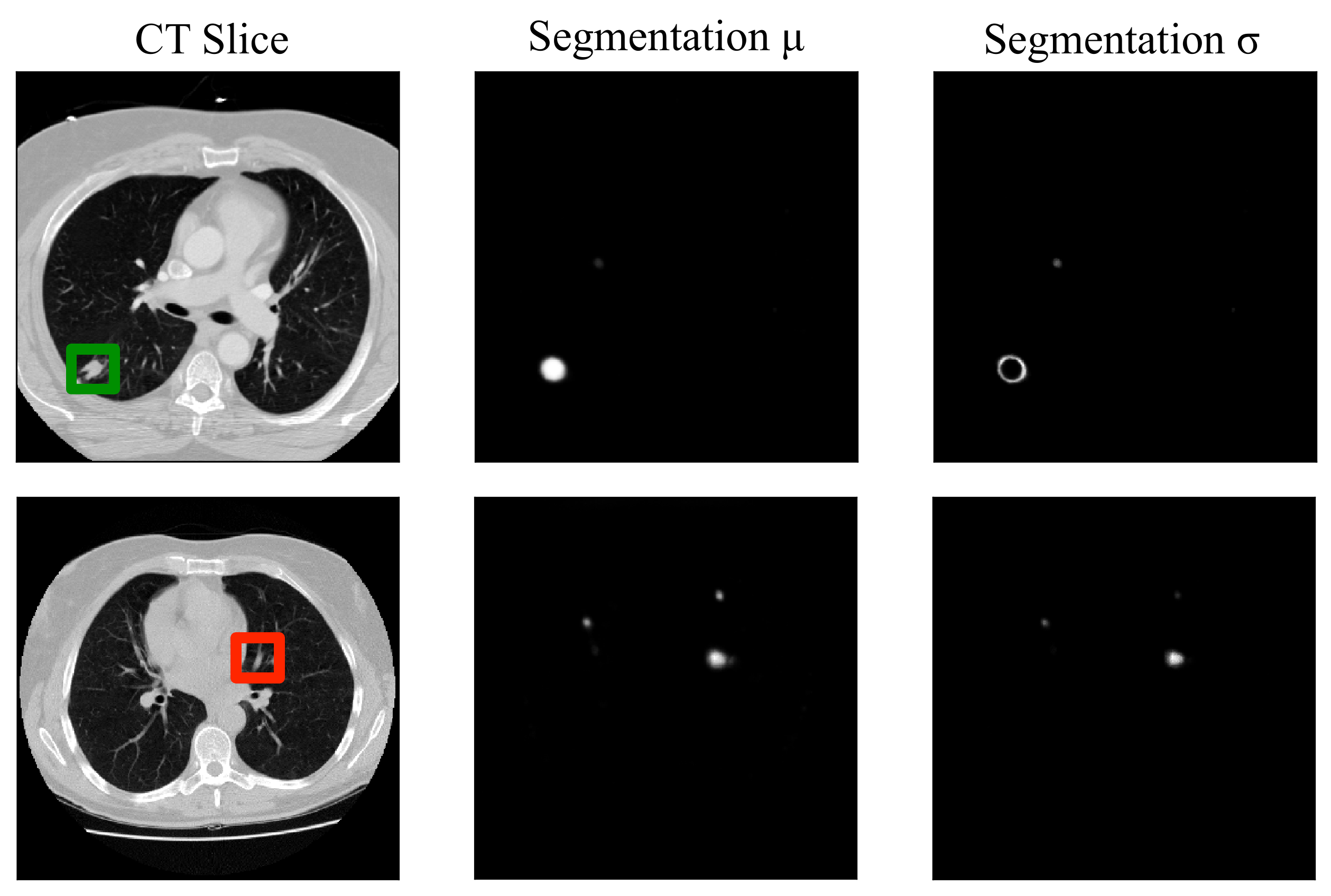}
		\caption{Full size examples}
		\label{fig:seg_ex_orig}
	\end{subfigure}
	~ 
	\begin{subfigure}[b]{0.49\textwidth}
		\includegraphics[width=\textwidth]{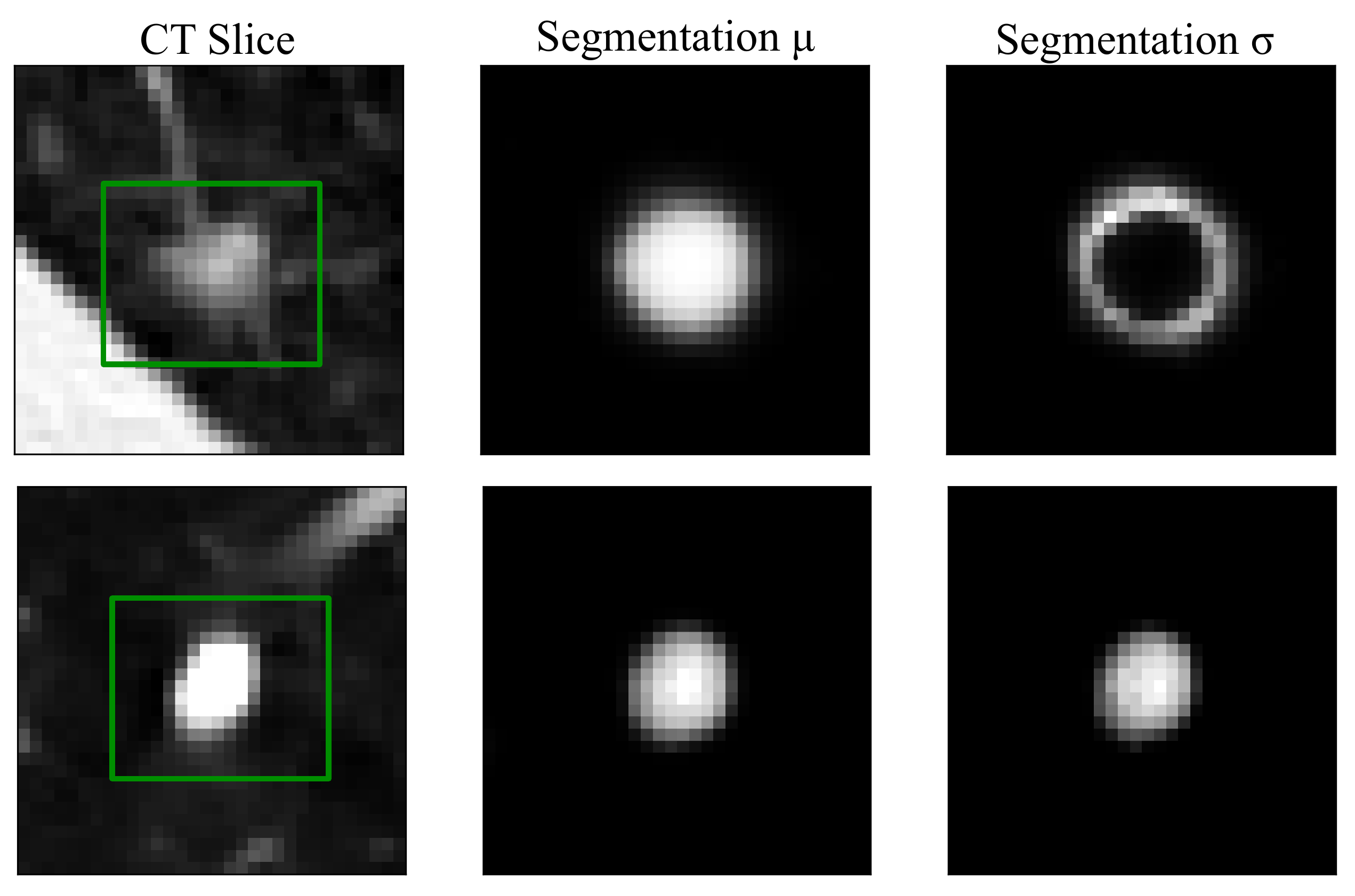}
		\caption{Zoomed examples (different than (a))}
		\label{fig:seg_ex_zoom}
	\end{subfigure}
	\vspace{-.25cm}
	~ 
	\caption{Example segmentation results. Left: Original axial CT slices. Green boxes and red box depict true positives and false positive after segmentation, respectively. Center: Segmentation mean $\mu$. Right: Segmentation uncertainty represented by standard deviation $\sigma$. The top row in (b) indicates an ideal case with confident segmentation predictions for true nodule location and uncertain predictions for the edges whereas the bottom row indicates that the model is overall uncertain.}\label{fig:seg_ex}\vspace{-.25cm}
\end{figure}

Since the segmentation network is Bayesian, we can approximate segmentation predictive mean ($\mu$) and standard deviation ($\sigma$) via MC samples as shown in Figure \ref{fig:seg_ex}. The top rows in Figure \ref{fig:seg_ex_orig} and \ref{fig:seg_ex_zoom} show example cases where the segmentation network is able to accurately identify true nodule pixels as well as most of the normal lung tissue with high confidence, with the exception of nodule borders where the predictive uncertainty is high. 
On images for which the segmentation network performs well with high confidence, these segmentation predictions help improve the performance of the 3D nodule detection network, especially since the segmentation network is able to see larger context in full 2D slices whereas the detection network has 3D local context. 

Segmentation prediction statistics are used as additional features for nodule detection by concatenating segmentation predictive mean and standard deviation with the original image to form a 3-channel composite image, as shown in Figure \ref{fig:arch}. We call this step `Uncertainty Fusion'. Our goal is to make nodule detection easier for cases such as shown in Figure \ref{fig:seg_ex} top row. 
Structural similarities between nodules and normal lung features in 2D means that the segmentation network that only uses 2D context can produce false positives, as well as lower confidence true positives.
We show such examples cases in the bottom row of Figure \ref{fig:seg_ex}. In Figure \ref{fig:seg_ex_orig} bottom row, there are false positives with high uncertainty, whereas in Figure \ref{fig:seg_ex_zoom} bottom row, there is a true detection with high uncertainty. To further improve performance, since there are difficult cases for which there is high overall uncertainty such as the ones shown in Figure \ref{fig:seg_ex_zoom} bottom row, we employ a second nodule detection network, trained using the original 1-channel image, without propagation of segmentation predictive uncertainty. As shown in the next section, combining predictions from these two detection networks enables us to combine the strengths of both approaches. 

\section{Experiments \& Discussion}\label{sec:res}

\begin{figure}[h!]
	\centering
	\begin{subfigure}[b]{0.49\textwidth}
		\includegraphics[width=\textwidth]{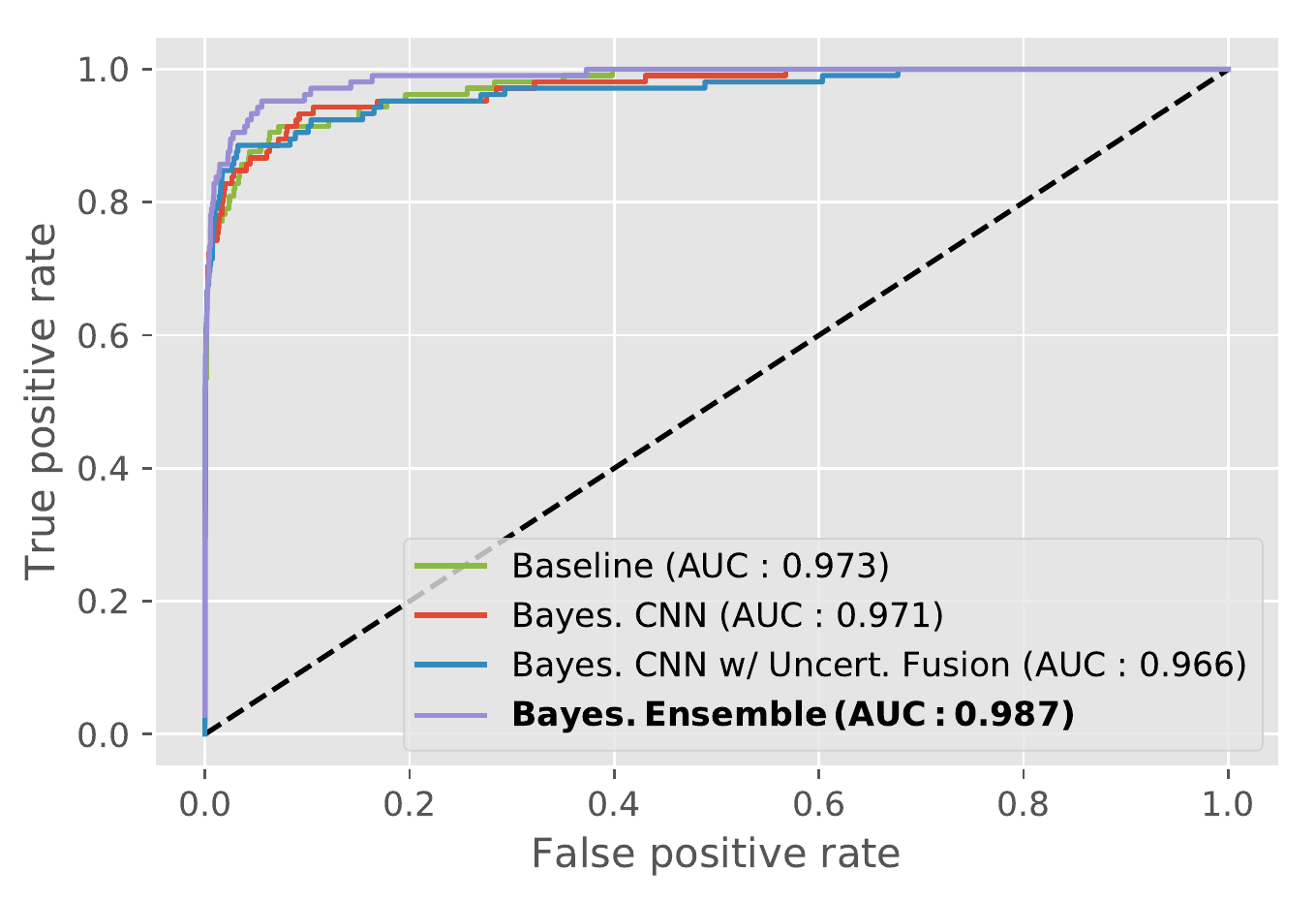}
		\caption{ROC Curves}
		\label{fig:roc}
	\end{subfigure}
	~ 
	\begin{subfigure}[b]{0.49\textwidth}
		\includegraphics[width=\textwidth]{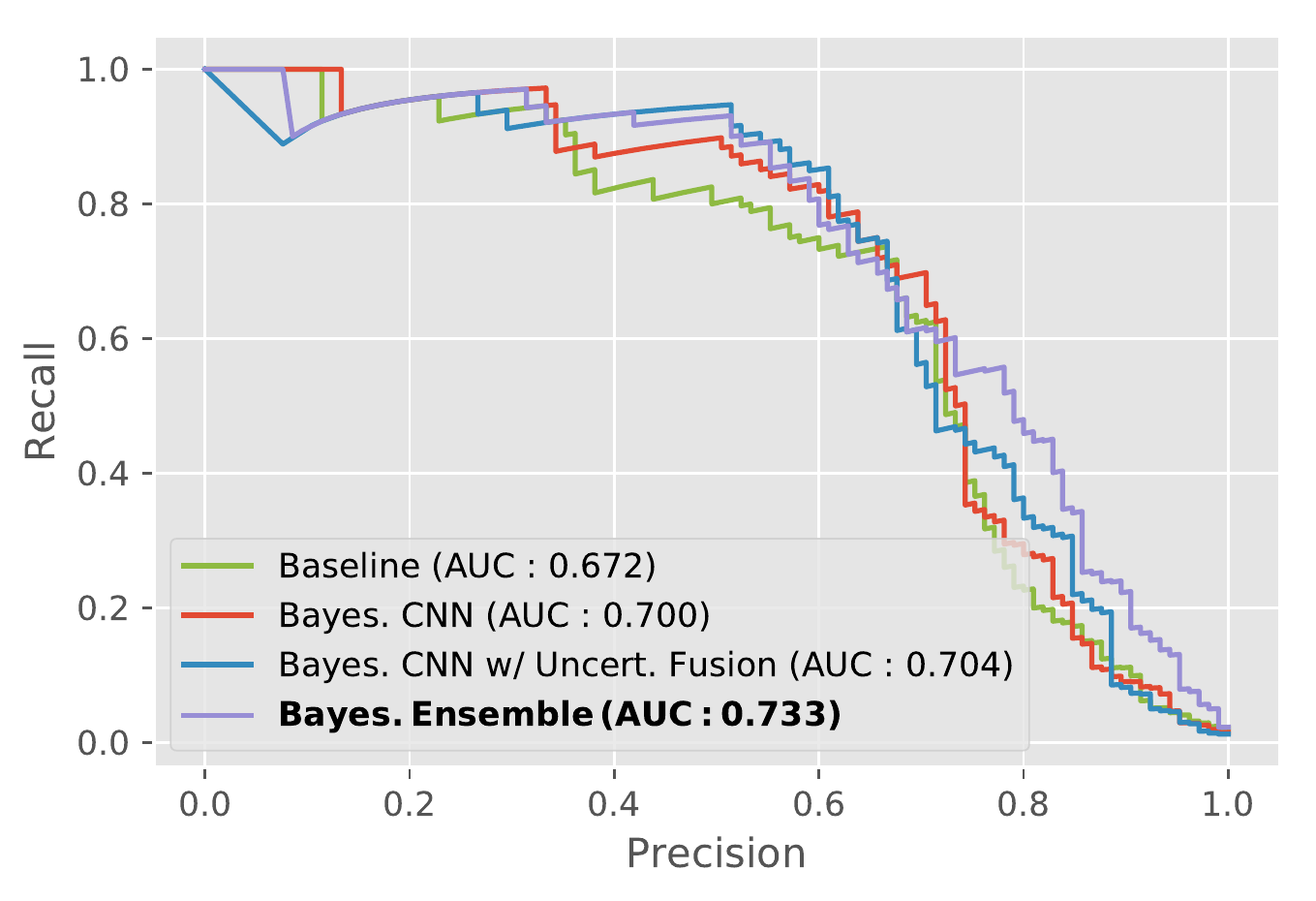}
		\caption{Precision-Recall Curves}
		\label{fig:prec}
	\end{subfigure}
	~ 
	\caption{3D nodule detection results and comparisons on test data for nodules extracted by the 2D segmentation network. }\label{fig:results}
\end{figure}

To evaluate the false positive reduction performance provided by the 3D nodule detection network, we use the LUNA16 dataset comprising 3D CT scans of 888 patients \cite{luna16}. Among 888 patients, 601 have nodules with a total number of 1186 nodules. We split the dataset into an $80\%/10\%/10\%$ train, validation, and test split. After segmentation, we get a recall and precision of ${\sim}92\%$ and ${\sim}0.9\%$ on the test data, respectively. 
Figure \ref{fig:results} shows the performance of Bayesian 3D nodule detection networks on the test data. Note that for these figures the total number of positives is limited to those that are extracted by the 2D segmentation network, as our goal is to compare the performance of the 3D nodule detection networks. If the segmentation network misses a nodule, that nodule can no longer be recovered by the detection network since the candidates for the detection network are generated based on stacked segmentation predictions. We present both ROC and Precision-Recall curves because the candidate set is extremely imbalanced (${\sim}0.9\%$ positive samples). As a performance benchmark, we also show results for a non-Bayesian 1-channel 3D CNN, called `Baseline'. We refer to the Bayesian nodule detection network that fuses uncertainty (via 3-channel composite image) as `Bayesian CNN with Uncertainty Fusion'. The Bayesian nodule detection network that only uses the original 1-channel image is referred to as `Bayesian CNN'. Both detection networks have exactly the same architectures. 

\begin{figure}[h!]
	\centering
	\includegraphics[width=.5\linewidth]{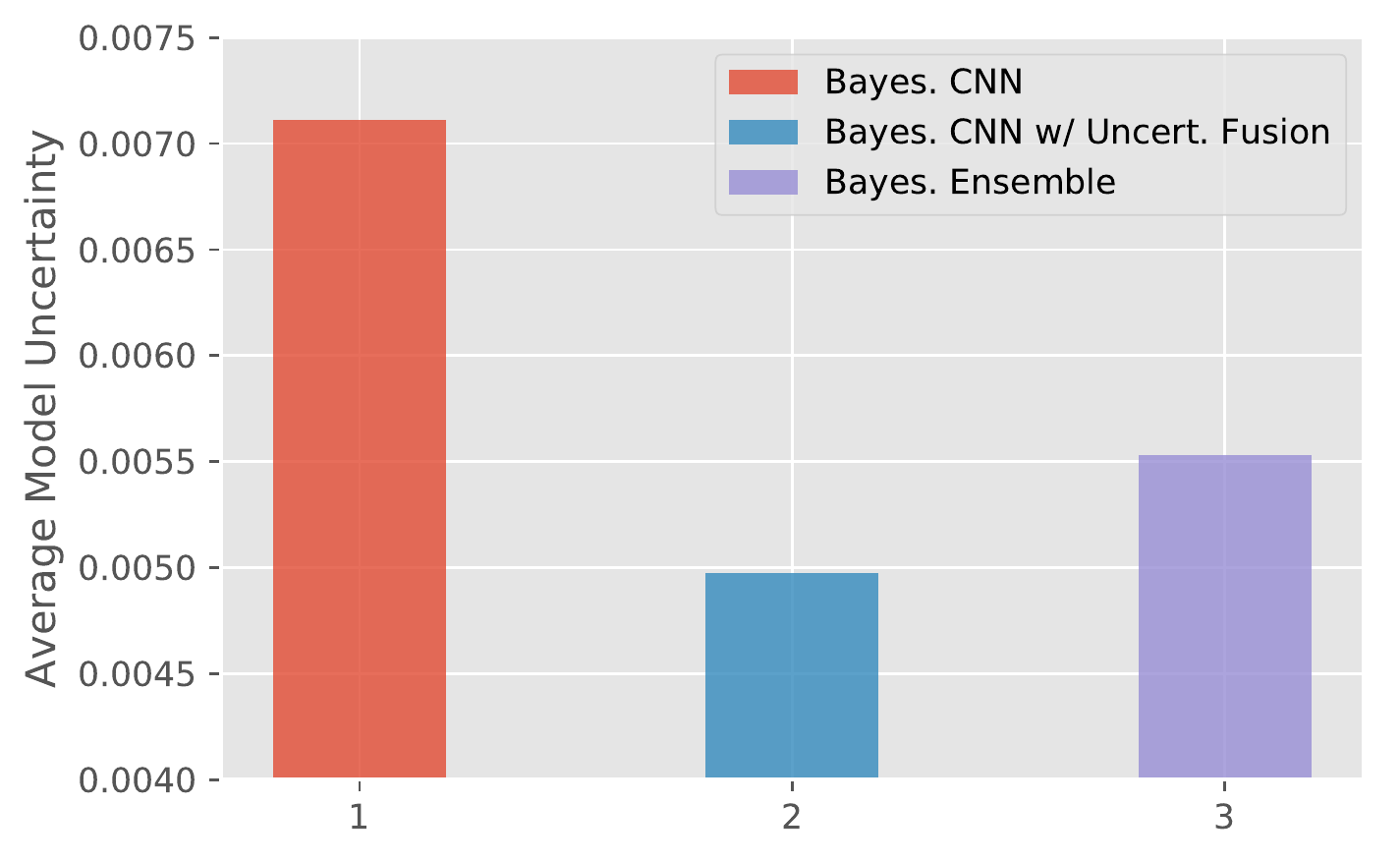}
	\caption{Average uncertainty of Bayesian CNN models. }\label{fig:avg_uncert}
	
\end{figure}

Figures \ref{fig:results} and \ref{fig:avg_uncert} show that the Bayesian CNN with Uncertainty Fusion has comparable AUC with the the 1-channel Bayesian CNN, but has higher overall confidence (lower average predictive uncertainty) due to the extra information provided by the segmentation network. We compute average predictive uncertainty in Figure \ref{fig:avg_uncert} via averaging standard deviations of predictions over all candidates. Example results on the validation set revealed that the Bayesian CNN with Uncertainty Fusion is able to make high confidence predictions for positive candidates similar to what is shown at the top row of Figure \ref{fig:seg_ex_zoom}, whereas it struggles for candidates similar to the bottom row. This is due to similar candidates that are negatives as shown at the bottom row of Figure \ref{fig:seg_ex_orig}. In contrast, the 1-channel 3D CNN struggles with some of the small nodules that the 3-channel 3D CNN easily identifies thanks to the uncertainty information provided by the segmentation network. 

To assess the dependence between predictions of the two Bayesian CNNs, we computed Spearman's rank corrrelation coefficient on the validation set resulting in $\rho=0.346$ ($p<10^{-5}$), which shows weak dependence between the predictions, confirming our manual inspection. Based on this information, we combined the two predictions via convex combination to create an ensembled prediction. We selected a weight of $0.5$ on each prediction, because it maximized the AUC on the validation set. As shown in Figure \ref{fig:results}, this ensembling approach resulted in significant improvement relative to either network operating independently. Further, denoting the test predictions of the two networks as $y_1^*$ and $y_2^*$, we can show that $y_1^*|x^*,\mathcal{D}$ and $y_2^*|x^*,\mathcal{D}$ are two independent random variables, where $x^*$ and $\mathcal{D}$ are the test input and the training data, respectively. Therefore, we can calculate the predictive variance of the ensembling approach for each ensemble prediction. 
Figure \ref{fig:avg_uncert} shows that the ensembling approach results in a slight increase in model uncertainty compared to the 3-channel Bayesian CNN (due to the large difference between the uncertainties of the two models), but this uncertainty is still much lower than that of the 1-channel Bayesian CNN. 

The performance of 3D Bayesian CNNs are comparable or better than the non-Bayesian baseline model. More importantly, unlike the non-Bayesian baseline, they provide model uncertainty/confidence information which could be extremely valuable for medical applications. We note that the performance improvement is more pronounced for the Precision-Recall curves since the positive class is extremely underrepresented in the data. 
We also show Brier scores \cite{deGroot83} (weighted due to heavy class imbalance) of test predictions in Table \ref{table:comp} 
showing that the Bayesian CNN with Uncertainty Fusion provides better performance than the 1-channel Bayesian CNN in terms of Brier score, and that ensembling the two methods provides the best performance.


\begin{table}[h]
	\small
	\caption{Brier Scores of 3D CNN Models}
	\label{table:comp}
	\centering
	\begin{tabular}{lc}
		\toprule
		Model     & Brier Score \\
		\midrule
		Baseline	&   0.1045   \\
		Bayes. CNN		&   0.1214   \\
		Bayes. CNN w/ Uncert. Fusion   	& 0.1010     \\
		Bayes. Ensemble     			  & \textbf{0.0948} \\
		\bottomrule
	\end{tabular}
\end{table}

\section{Conclusion}\label{sec:res}
Focusing on a CAD problem, we proposed a method to propagate and fuse uncertainty information in a multi-stage Bayesian convolutional neural network (CNN) architecture. Our experiments showed that propagating and fusing uncertainty improved the overall performance in terms of both final prediction accuracy and model confidence. Although the specific problem we considered in this paper was CAD, the proposed method could be used in a variety of applications for which the knowledge of uncertainty is important and informs subsequent decision making processes in the overall system. Example applications include medical image based diagnostics, autonomous driving, and robotics. 

Our results pave the way for several interesting research questions that we will investigate as future work. Although we performed ensembling to combine the predictions from two detection networks, we will explore whether we can integrate this ensembling process within a single CNN architecture and have a single network learn how to best combine the information from different channels. 
We will also consider adaptive ensembling approaches, where predictions are combined by using the information about each prediction's uncertainty. In addition, we will investigate other approximate Bayesian inference techniques such as variational dropout \cite{kingma_nips15} and assess how close the estimated uncertainties via different approximations are to true uncertainties. This information could be crucial for certain applications such as medicine, where overestimation of uncertainty may be preferable to underestimation. Finally, we will investigate whether we can separately model the uncertainty due to lack of training data and the uncertainty due to inherent noise in the data \cite{gal_arxiv17}, which may have different implications for subsequent decision making processes.

\bibliographystyle{unsrt}
\bibliography{refs}
\end{document}